# Research on Optimization Method of Multi-scale Fish Target Fast Detection Network

Yang Liu[1, 2], Shengmao Zhang[2], Fei Wang[2], Wei Fan[2], Guohua Zou[3], Jing Bo[4]

（1 School of Information, Shanghai Ocean University, Shanghai, 201306; 2 Key Laboratory of East China Sea Fishery Resources Exploitation, Ministry of Agriculture; East China Sea Fisheries Research Institute, Chinese Academy of Fishery Sciences, Shanghai, 200090; 3 Shanghai Junding Fishery Science and Technology Co. Ltd., Shanghai, 200090; 4 Southwest Forestry University, Kunming City, Yunnan Province，650233)

**Abstract**：The fish target detection algorithm lacks a good quality data set, and the algorithm achieves real-time detection with lower power consumption on embedded devices, and it is difficult to balance the calculation speed and identification ability. To this end, this paper collected and annotated a data set named "Aquarium Fish" of 84 fishes containing 10042 images, and based on this data set, proposed a multi-scale input fast fish target detection network (BTP-yoloV3) and its optimization method. The experiment uses Depthwise convolution to redesign the backbone of the yoloV4 network, which reduces the amount of calculation by 94.1%, and the test accuracy is 92.34%. Then, the training model is enhanced with MixUp, CutMix, and mosaic to increase the test accuracy by 1.27%; Finally, use the mish, swish, and ELU activation functions to increase the test accuracy by 0.76%. As a result, the accuracy of testing the network with 2000 fish images reached 94.37%, and the computational complexity of the network BFLOPS was only 5.47. Comparing the YoloV3~4, MobileNetV2-yoloV3, and YoloV3-tiny networks of migration learning on this data set. The results show that BTP-Yolov3 has smaller model parameters, faster calculation speed, and lower energy consumption during operation while ensuring the calculation accuracy. It provides a certain reference value for the practical application of neural network.

Fund: National Natural Science Foundation of China under Grant No. 61936014; Open project of Key Laboratory for sustainable utilization of marine fishery resources in Zhejiang Province under Grant No.2020KF001;National Natural Science Foundation of China under Grant No. 31772899
Author: Liu Yang (1996—), graduate student, research direction: fish image analysis, computer vision and other fields.
E-mail:yangzai126@126.com
Corresponding author: (1976 -) Zhangsheng Mao, Ph.D., associate professor, research direction: fish image analysis. E-mail: ryshengmao@126.com

# 1. Introduction

The aquarium market for ornamental fish is a multi-billion dollar industry, and aquarium farming is an important way to conserve wild fish, spread marine knowledge, and provide an important source of income for some people. However, China's artificial breeding technology lags behind that of developed countries [1]. In the breeding process of ornamental fish, the detection and tracking of fish behavior will be helpful to monitor the health status of fish. In the process of tourists watching, the detection and identification of fish can better disseminate Marine knowledge [2].

In recent years, deep learning is widely used in the field of target detection. The existing target detection networks include regional suggestion detection network and end-to-end detection network. The regional suggestion network includes Faster R-CNN [3],Fast R-CNN [4], R-FCN [5] and Libra R-CNN [6], but it is difficult for the regional suggestion network to achieve real-time performance. End-to-end networks can achieve real-time detection, including RPN [5], YOLO [7, 8, 9], SSD [10] and Retinanet [11]. There are two main methods to apply the existing network to the more subdivided field of fish target identification. The first method is to retrain the existing network with a larger fish data set, and the second method is to transfer the existing network with a smaller fish data set. For example, Hongchun Yuan and others [12] migrated and learned Fast Rcnn network, and the average test accuracy reached 91.7%. However, the backbone, which uses VGG[30] and Resnet101 as models, has large model parameters. Li Qingzhong et al. [13] modified Yolo's backbone to avoid revealing the large parameters of the network model, and achieved an accuracy of close to 99% on 1500 images, but the data was too small. Zhang Shengmao et al. [2] retrained the SSD-Mobilenet V1 network and achieved an identification accuracy of 92.5%, but did not further optimize the algorithm.

In recent years, the optimization of target detection and identification algorithm has been the focus of research, mainly including data augmentation and network structure optimization.The data augmentation includes image stretching, translation, flipping, rotation and cropping, Gaussian noise, color transformation, Mixup[14], Cutmix[15], Mosaic[16], contrast learning generation[17], style transfer and Autoaugment[18]. Network structure optimization includes Residual Block proposed by K. He et al. [19] to prevent gradient disappearance; Depthwise proposed by Chollet F et al. [20] to reduce the amount of parameters; the Pointwise module proposed by Ioffe et al. [21] to flexibly change the channel; Inverted Residuals and Linear Bottlenecks proposed by Sandler et al. [22] on the basis of [20, 21] to prevent feature degradation. In recent years, the most commonly used activation functions for target detection algorithms include the non-smooth functions ReLU[23] and Leaky ReLU[24] with less calculations, and the smooth functions ELU[25], Swish[26], Elish [27] and Mish [28] with more calculations.

In conclusion, the fish target detection algorithm needs more annotated data sets in order to make models with stronger generalization ability and higher identification accuracy. In addition, due to the existing network training, retraining or transfer learning based on public data sets such as ImageNet data set, MS COCO data set or Pascal VOC data set, the parameters is large. Finally, the identification ability of fish target detection network can be further enhanced by network design, data augmentation and activation function debugging. Based on this, our contributions are summarized as follows:

(1) We collect, clean, and annotate a dataset named "Aquarium Fish", consisting of 84 common fish with a total of 10,042 images

(2) We propose a multi-scale fast fish target detection network (BTP-yoloV3). Compared with Yolov4[16], the parameters decreases by 94.1%, the mAP [34] (Mean Average Precision) value on this data set reaches 94.37%, the network size is only 13.9M, and the BFLOPS (Billion float operations) is only 5.47.While the accuracy is maintained on the data set, the identification speed is faster.

(3) We use different data augmentation and activation functions to optimize the

network, which increases mAP by 3.32% in total.

## 2. Data and methods

First, the experiment is inspired by the design ideas of Dw, Pw[20], BN[29], IR and LB[12] modules, and redesigns the BTP-Yolov3 network, which has the advantages of fast identification speed and multi-scale detection. Then, in order to improve the generalization ability of the network, we use Mixup[14], Cutmix[15] and Mosaic[16] to increase the training of data sets, and use activation functions such as ELU[25],Swish[26] and Mish[28] to transfer and learn the network and optimize the ability of network identification and detection. Finally, the experiment is compared with Yolov4[16], Yolov3[9], Yolov3-tiny[9], and Mobilev2_Yolov3 to transfer learning on this data, proving that the network can achieve the fastest detection speed while maintaining the identification ability.

### 2.1 The Data Processing

Large-scale fish data can facilitate the development of more powerful and complex identification networks and algorithms. However, as fish species are diverse, how to accurately organize and collect data sets is still a key issue. For this purpose, we collect fish video images in the aquarium and manually cleaned to remove ambiguous and non-fish images. Finally, we obtained 10042 images containing 84 species of fish data, and the resolution of each image was (600×400) pixels, as shown in Fig. 1 (a). In order to meet the requirements of target detection and identification based on supervised learning of this data set, we use Labelme software to label the images, and each image includes target bounding box (x, y, w, h) and category label information. x, y, w and h respectively represent the position (x, y) and length and width (w, h) of the bounding box. In the experiment, the data were summarized and sorted according to the format of VOC(Visual Object Classes) to facilitate reuse by others, as shown in Figure 1 (b).

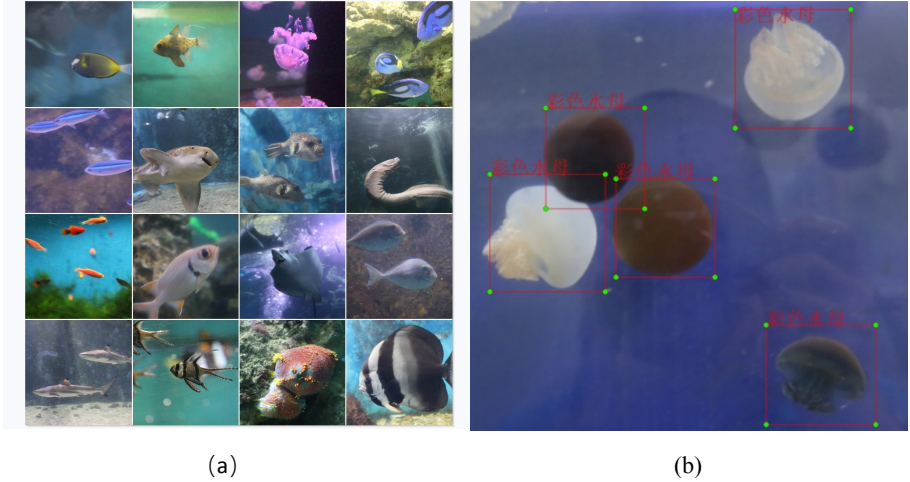

(a)          (b)

Fig.1 (a) Sample Of Marine Life Data In Aquarium, (b) Pascalvoc Format Image Annotation Of Color Jellyfish In Aquarium

### 2.2 Data Augmentation

The aquarium fish data set was divided into test set training set and test set according to the ratio of 8:2. A large number of diverse training sets can augment the generalization ability of the network. Therefore, to augment the image data, the experiment first stretches, translates, flips, rotates and crops the image. Then the image was enhanced with Mixup[14], Cutmix[15], and Mosaic[16], as shown in Figure 2.

Mixup is based on the priori principle that linear interpolation of feature tensor corresponds to linear interpolation of related labels, with low computational overhead and data independent [14], as shown in Equation 1.

$$\begin{cases}\tilde{x} = \lambda x_i + (1-\lambda)x_j \\ \tilde{y} = \lambda y_i + (1-\lambda)y_j\end{cases} \#(1)$$

In the formula, $x_i, x_j$ represent two different input tensors, $y_i, y_j$ represent the corresponding One-Hot labels, and λ belongs to the Beta distribution.

Cutmix cuts a piece of image A and fills it with B, which increases the complexity of the data in the reasoning process and improves the generalization ability of the model [15], as shown in formula 2.

$$\begin{cases} \tilde{x} = M \odot x_A + (1-M) \odot x_B \\ \tilde{y} = \lambda y_A + (1-\lambda) y_B \end{cases} \#(2)$$

In the formula, $x_A, x_B$ represent two different training samples, $y_A, y_B$ represent the corresponding One-Hot label, $M$ represents the filled mask matrix (the filled area is 1, the others are 0), and λ belongs to the Beta distribution.

Mosaic combines four images into one image through clipping, which increases the data complexity on the original data and makes the model have a better robustness [16].

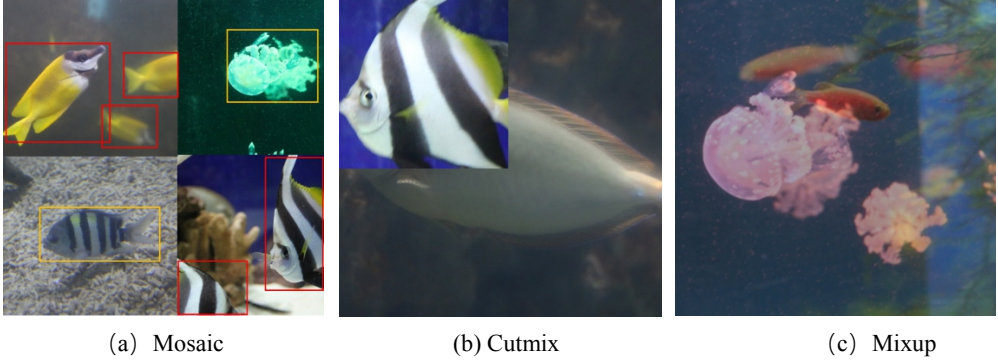

(a) Mosaic      (b) Cutmix      (c) Mixup

Fig. 2 (a) Mosaic Data Enhancement, (b) Cutmix Data Enhancement, (c) Mixup Data Enhancement.

**2.3 Aquarium Fish Target Detection Network**

The available backbones for target detection algorithms include VGG[30], Resnet [19], Densenet[31], Squeezenet[32], MobilenetV1~2[22,35], etc. Using them to transfer and learn aquarium fish data may cause parameter redundancy and slow real-time detection. The main reason is that CNN convolution requires a large amount of calculation and has many network layers. Inspired by MobilenetV1~2[22,35] and Xception[20] to reduce the amount of calculation by using depthwise separable convolution, we redesigns the target detection network backbone module, as shown in Figure 3,4. The module is mainly composed of PDP1-3 (Pointwise To Depthwise To Pointwise) using the residual principle, so that its feature information can be transmitted stably; In order to reduce the amount of calculation, this module uses Depthwise convolution (convolution kernel is $K^2$) to calculate the characteristic tensor of (W, H, N), the calculation amount is W×H×$K^2$, and it is only 1/N of the calculation amount of CNN convolution W×H×N×$K^2$, and its identification accuracy does not affect too much [20]; In order to compress and expand tensor information flexibly, Pointise convolution is used to fuse tensor information after Depthwise convolution, and Linear activation function is used to prevent feature degradation in the shallow layer network [22]; In order to speed up training and prevent overfitting, we use the Batch Normalization(BN) module after convolution [29]; The Mish activation function is used to make the tensor transition smoother in the reverse and forward propagation process; Finally, the maximum pooling with a step size and a size of 2 is used for downsampling. Suppose the input tensor is (W, H, N). Firstly, the feature information is extracted from N channel through PDP1, and the output tensor is (W, H, n)；Secondly, PDP2 extracts fine-grained information. PW convolution is used to expand the feature channel to 2n to refine tensor information, then Dw is used to extract fine-grained features. Finally, Pw compresses information back to N-channel and outputs tensors [22] (W, H, n); Thirdly, PDP3 takes the residual tensor (W, H, 2n) combined by PDP1 and PDP2 on the channel to retain more original information and prevent feature degradation in the process of PDP2 extraction [13], and uses Pw and Linear to compress information and output the tensor (W, H, n); In the end, the residual tensors of PDP1 and PDP3 were spliced together, and 2×2 maximum pooling (maxpool 2:2) was used for down-sampling. For convenience, this restructured network is referred to as TPDP(Three PDP) in this paper.

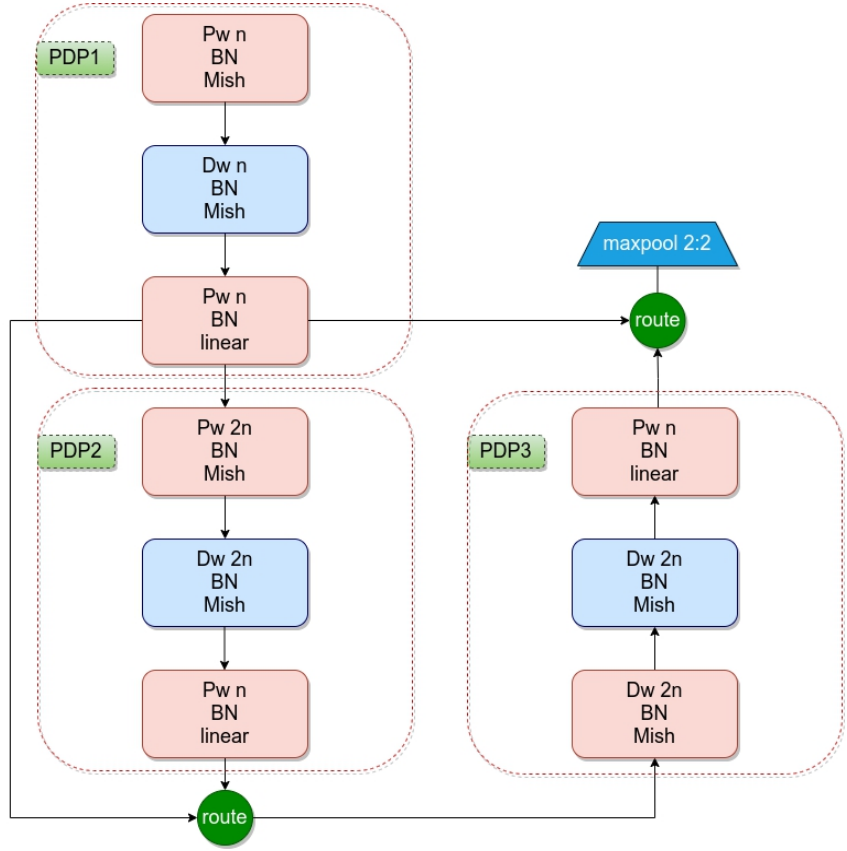

Fig.3 reorganized convolution module (TPDP). dw n means that the convolution kernel is (3×3) depthwise convolution of channel n, pw means that the convolution kernel is (1×1) pointwise convolution of channel n, and bn means batch normalization.

The experiment uses three TPDP serial connections to form a backbone, among which the channels of PDP1~3 are 32, 64, and 128 in sequence. This article abbreviates the backbone as BTP (The backbone by TPDP). BTP achieved the highest mAP value of 94.37% in the experiment. As shown in Fig. 4 (A), the modules of this network include backbone, neck and head .In this backbone, we firstly use 32-channel convolution with the convolution kernel size of (3×3) and step size of 2 to initially extract image features. Then we use three TPDP to compress, extract and fuse the features for several times. In neck, low-level feature semantic information and high-level feature semantic information are fused into feature pyramid for hierarchical prediction, which can better describe small targets [33]; The output of TPDP2 and TPDP3 were fused into two branches in the neck. In the first branch, the feature tensor of TPDP3 output (16,16,512) was compressed into a new feature vector (16,16,256) by a Pointwise, and then the conventional convolution kernel (3×3) was used to reach the part of head1. In the second branch, the fusion of low-level feature information and high-level feature information can enlarge the receptive field. The feature tensor (16, 16, 512) output of TPDP3 at a high level is firstly convolved by CNN, then compressed into (16, 16, 128) by a Pointwise, and then expanded to (32, 32, 128) by an upsampling module; Finally, the low-level feature vector (32×32×128) and high-level feature vector (32×32×128) of PDP2 output are fused together to reach head2.We use 32 times of head1 downsampling to detect small targets, and 16 times of head2 downsampling to detect large targets. And because the tensor divides 2 times of downsampling in length and width, the output features of head1 and head2 can be compatible with multi-scale image input. The experiment uses the design of Yolov3[9] to divide the picture into S×S grid units, each grid unit predicts 3 bounding boxes, each bounding box contains 5 parameters $(x, y, w, h, \eta)$, which respectively represent the position coordinate *(x, y)* of the prediction bounding box, the width and height of the prediction bounding box *(w, h)* and the target confidence of the grid *(η)*.In this experiment, there are a total of 84 species of fish, so the channel of the convolution module in head1 and head2 is 267 (number of bounding boxes ×(number of categories + number of bounding box parameters) : 3×(84+5)), as shown in Fig. 4 (b). For convenience, this paper calls the network BTP-YoloV3.

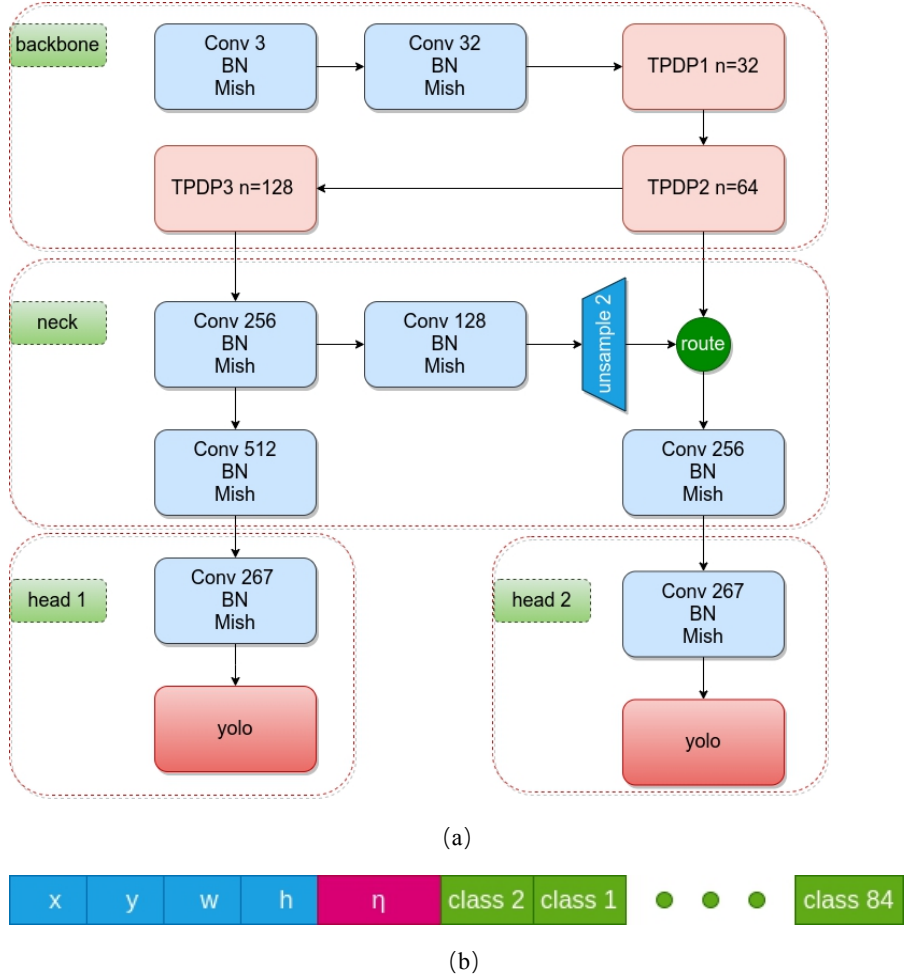

Figure 4  (a) network structure, (b) predicted parameters of each detection frame

## 2.4 Evaluation Function

The loss functions of the experiment are classification loss, target classification loss and target confidence loss. After dividing the image into S2 grids, the network hopes to evaluate whether the target is included in each grid. Target detection is performed if the target is included, and the next grid is scanned if it is not. The confidence loss function adopts the cross entropy function, as shown in Formula (3), $S^2$ represents the number of segmented grids; $M$ represents the number of bounding boxes; $\eta_i$ and $\hat{\eta}_i$ represent the real and predicted values of the grid. $1_{ij}^{obj}$ means 1 if there is a target in the $j^{th}$ prediction bounding box of the $i^{th}$ grid, or 0 if there is a target in the $j^{th}$ prediction bounding box; $1_{ij}^{noobj}$ means that if the $j^{th}$ prediction bounding box in the $i^{th}$ grid does not contain a target, it is 1, otherwise it is 0.

The positioning loss is the same as CIOU [37]. The gap between the real box and the prediction bounding box is the proportion of the overlap area between them and the sum of their areas. The overlap area between them is expressed as G (Formula 4), $\hat{x}, \hat{y}, \hat{w}$ and $\hat{h}$ represent the abscissa, ordinate, width and height of the prediction bounding box respectively. x, y, w, and h represent the abscissa, ordinate, width, and height of the real box respectively.

$$L_{cls} = -\sum_{i=0}^{S^2}\sum_{j=0}^{M} 1_{ij}^{obj} [\hat{\eta}_i \log(\eta_i) + (1-\hat{\eta}_i)\log(1-\eta_i)]$$
$$-\sum_{i=0}^{S^2}\sum_{j=0}^{M} 1_{ij}^{noobj} [\hat{\eta}_i \log(\eta_i) + (1-\hat{\eta}_i)\log(1-\eta_i)] \#(3)$$

$$G = \left[\text{Min}(\hat{x}+\hat{w}, x+w) - \text{Max}(x,\hat{x})\right] \times \left[\text{Min}(\hat{y}+\hat{h}, y+h) - \text{Max}(y,\hat{y})\right] \#(4)$$

IoU[36] represents the gap between the real box and the prediction bounding box, and the more overlapping areas, the greater the value of IoU [36]. Formula (5).

$$IoU = \frac{G}{w \times h + \hat{w} \times \hat{h} - G} \#(5)$$

In addition, the gap between the aspect ratio of the real box and the prediction bounding box can narrow the distance between them and measure the consistency of the aspect ratio of the real box and the prediction bounding box, as shown in Formula (6).

$$v = \frac{4}{\pi^2}(\arctan\frac{w}{h} - \arctan\frac{\hat{w}}{\hat{h}})^2 \#(6)$$

The overlapping case should have more priority than the non-overlapping case. Alpha can make the overlapping case take precedence over the non-overlapping case (Formula 7).

$$\alpha = \frac{v}{1 - IoU + v} \#(7)$$

In order to solve the shortcoming that IoU cannot evaluate the overlapping area, we use the Euclidean distance of the center point of the prediction bounding box and the real box to calculate, which can not only evaluate the distance of the non-overlapping area, but also evaluate the distance of the overlapping area. Finally, the positioning loss function is shown in formula 8.

$$L_{CIoU} = 1 - IoU + \frac{\|b,\hat{b}\|_2^2}{c^2} + \alpha v \#(8)$$

In formula (8), b and $\hat{b}$ represent the coordinate centers of the real box and the prediction bounding box respectively.

In the classification loss, cross entropy loss function is used to evaluate the distance between the predictions category and the real category, as shown in formula (9). *C* represents category and *Class* represents category set. $\hat{p}_i(c)$ represents the prediction value and $p_i(c)$ represents the real value.

Finally, the loss function is formula (10), $\lambda_{cls}$ and $\lambda_{CIoU}$ are pre-variables between 0 and 1, which are set to 1 in this paper.

$$L_c = -\sum_{i=0}^{S^2} 1_{ij}^{obj} \sum_{c \in classes} [\hat{p}_i(c)\text{Log}(p_i(c)) + (1-\hat{p}_i(c))\text{Log}(1-p_i(c))] \#(9)$$

$$L = \lambda_{cls}L_{cls} + \lambda_{CIoU}L_{CIoU} + L_c \#(10)$$

## 3. The Experiment

The experiment uses the Darknet deep learning framework to run on a Tesla V100 graphics card with 16G of GPUmemory and a Linux system with 32G of memory.

### 3.1 The Evaluation Index

The experiment uses BFLOPS (Billion Float Operations), Average IoU, mAP[34] and FPS to evaluate the algorithm complexity, positioning accuracy, average test accuracy and the number of images calculated per second of network backbone.The calculation formula of BFLOPS is as follows:

$$\sum_{c=0}^{C} 2W \cdot H \cdot N \cdot M \cdot K^2$$
$$+ \sum_{d=0}^{D} 2w \cdot H \cdot K^2 + 2M \cdot N \quad \#(11)$$

In formula (11), W and H represent the length and width of the input feature, N and M represent the length and width of the output feature, K represent the size of the convolution kernel, C and D respectively represent the number of convolution of conventional CNN convolution and Depthwise-pointwise convolution. Average IoU is to average the verification IoU. mAP is the average accuracy index.

### 3.2 BTP-Yolov3 Results

The network uses the SGD optimizer to update the network weights during back propagation, with initial learning rate of 0.0013, momentum of 0.949, decay of 0.0005, and initial input image of 512×512. The Batch Size is 512, the video memory is loaded in 16 times to calculate forward propagation, and then to calculate the back propagation. In addition, the learning rate is reduced to 1/10 of the original one when iterating 20,000 and 40,000 Batches until complete convergence. In the experiment, the image was randomly expanded, translated, flipped, rotated and cropped in space, which was taken as the basic data augmentation steps. Mixup[14], Cutmix[15] and Mosaic[16] were used for further data augmentation; In order to speed up network training, we firstly replaces the Mish activation function with the ReLU activation function for warm-up training of the network. After the weight stabilizes, the Mish activation function is used to transfer the learning network with the same super parameters to enhance the robustness of the network. Finally, the mAP of the test set of 2000 images reaches 94.37% in the experiment, as shown in Figure 5.

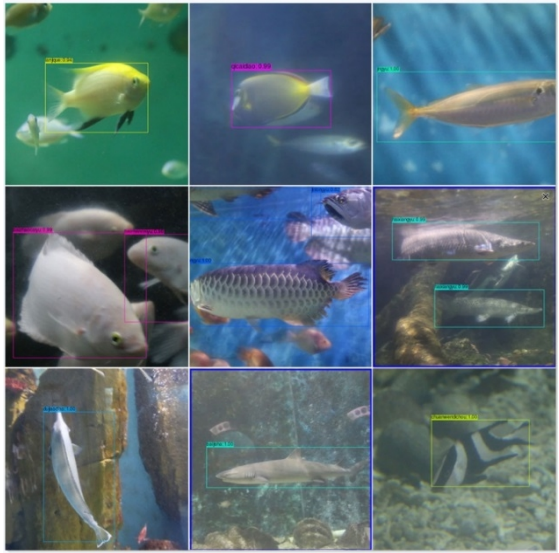

Fig.5 fish target detection results

### 3.3 Multi-scale Image Input

The training image has a large amount of data, so we scale, Mosaic and Cutmix the image. And the network has generalization ability for different sizes of the same object. In addition, head1 and head2 downsample images by 32 times and 16 times respectively, and the output tensor dimensions are (16, 16, 256) and (32, 32, 256), which can be compatible with the expansion and contraction of 2 times images as the input data of the network to a certain extent. After testing, the network is effectively compatible with 6 image sizes after training on 512×512 images, as shown in Table 1. Since the network is trained on 512×512 images, its mAP and Average IoU perform best. As the image size decreases, its index also declines. The network realizes the compatibility of image output size.

Tab. 1 Multi-Scale Image Index

| Image Size | mAP | Average IoU |
| --- | --- | --- |
| 608×608 | 92.57 % | 63.71 % |
| 512×512 | **94.37%** | **72.81%** |
| 416×416 | 94.23 % | 69.83 % |
| 306×306 | 88.16 % | 63.67 % |
| 256×256 | 85.26 % | 66.13 % |
| 208×208 | 60.62 % | 54.33 % |

### 3.4 BTP-Yolov3 Network Optimization

In order to quantitatively study the influence of network tuning techniques on the evaluation index, experiments respectively verified the influence of different data enhancement and activation functions on mAP. A total of 9 experiments were carried out, as shown in Table 2. The experimental results show that the data-augmentation Mixup, Cutmix and Mosaic methods all improve the network identification ability, and the Mosaic data augmentation improves more. Compared with non-smooth ReLU activation functions, smooth activation functions like Swish, Mish and Elu can get better results, and Mish fits the network best. Using BN can accelerate the fitting speed and improve the robustness of the network.

Tab.2 The Effect of Different Data Enhancement

| Mixup | Cutmix | Mosaic | ReLU | Swish | Mish | ELU | BN | mAP@0.5 |
| --- | --- | --- | --- | --- | --- | --- | --- | --- |
|  |  |  | ☑ |  |  |  | ☑ | 92.34% |
| ☑ |  |  | ☑ |  |  |  | ☑ | 92.69% |
|  | ☑ |  | ☑ |  |  |  | ☑ | 92.93% |
|  |  | ☑ | ☑ |  |  |  | ☑ | 93.12% |
| ☑ | ☑ | ☑ | ☑ |  |  |  | ☑ | **93.61%** |
| ☑ | ☑ | ☑ |  | ☑ |  |  | ☑ | 93.01 % |
| ☑ | ☑ | ☑ |  |  | ☑ |  | ☑ | **94.37%** |
| ☑ | ☑ | ☑ |  |  | ☑ |  |  | 89.66% |
| ☑ | ☑ | ☑ |  |  |  | ☑ | ☑ | 93.27% |

### 3.5 Compare with Other Networks

In the field of target detection, backbones that are used more include VGG [30], Resnet [19], Densenet [31] and MobilenetV1~2[22,35]. VGG, Resnet, and Densenet have a large amount of parameters, and the calculation speed of the training model is slow, and real-time detection cannot be achieved. So they are not suitable for this experiment. Mobilenet uses Depthwise convolution, which greatly reduces the amount of network calculation and has a faster running speed on embedded devices. Therefore, the experiment combines Mobilenetv2 [22] with the head of Yolov3 to form Mobilenetv2-Yolov3 to train on this data as the control group. Since Yolov3[9], Yolov4[16], and Yolov3-tiny can achieve real-time detection, the experiment uses their pre-training weight transfer learning as a control group. In the training process of the network, 2000 test images are used, and the mAP and Recall indicators are tested every 1000 times of backpropagation, as shown in Fig.6. Due to the amount of parameters in Yolov4[16], mAP and Recall oscillate severely. The Mobilenetv2-Yolov3 model is relatively less oscillating.BTP-Yolov3 and Yolov3-Tiny fitted stably, and BTP-Yolov3 fitted the fastest and had a better performance. It is proved that using BN can accelerate the fitting speed of the model and improve the generalization ability.

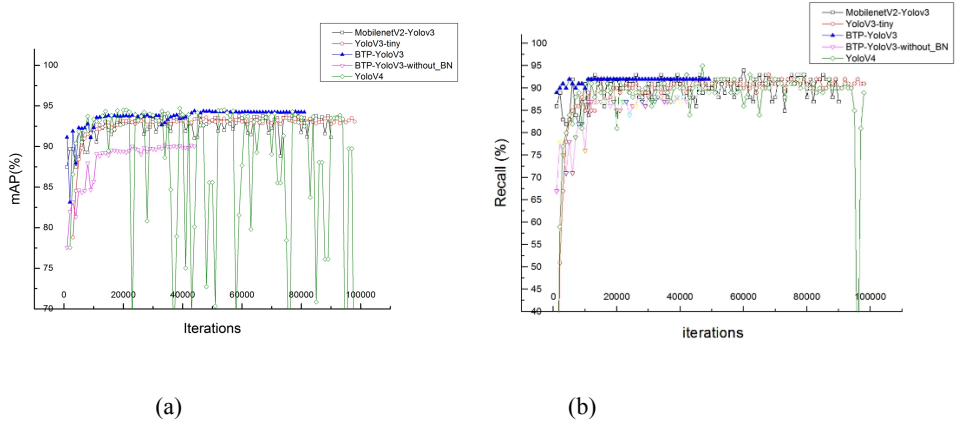

Fig. 6 accuracy and recall rate during training. (a) average test accuracy, (b) recall rate.

In order to comprehensively evaluate the indexes of each model, the experiment sets the Batch Size of all models to 1, and the inputs image size to 512×512, and tests on the idle Tesla V100 graphics card, as shown in Table 3. Yolov4[16] achieved the highest mAP on this data, but its algorithm complexity (BFLOPS) is too large, the weight is large, the runtime memory usage is high, and the detection speed (FPS) is slow. Yolov3-tiny[9], Mobilenetv2-Yolov3 and BTP-Yolov3 all have the characteristics of small weight, high test accuracy and high running speed. BTP-Yolo has the highest detection speed (FPS), and the mAP is higher than others in small models, achieving a better balance between test accuracy and speed.

Tab. 3 Comparison of other Network Evaluation Indexes

| Network | Weight Size | Average_IoU | mAP | BFLOPS | FPS |
| --- | --- | --- | --- | --- | --- |
| Yolov3 | 237M | 69.58 % | 93.68% | 99.84 | 105.3 |
| Yolov3-tiny | 34M | 66.41 % | 93.34 % | 5.577 | 250 |
| Yolov4 | 258M | 70.75% | **95.40%** | 91.13 | 87.0 |
| Mobilenetv2-Yolov3 | **7.7M** | 71.94 % | 93.42 % | **4.786** | 222.2 |
| Mobilenetv1-Ssd | 33.8M | - | 92.5% | - | - |
| BTP-Yolov3 | 13.9M | **72.81%** | 94.37% | 5.476 | **285.7** |

## 4. Conclusions

The experiment collected, cleaned, and annotated the aquarium fish data set, and proposed a multi-scale input fish target fast detection network whose mAP value reached 94.37%, and the network size was only 13.9M; Compared with traditional networks (Yolov3~4, Mobilenetv2-Yolov3, Yolov3-tiny), this network has achieved a better balance of speed and accuracy on this data set. In the case of improving the detection speed, we have prevented the weakening of identification and detection capabilities, which can be better applied to embedded systems and provide some references for the optimization of neural networks in applications.